\documentclass{article}
\usepackage{ijcai25}

\usepackage{times}
\usepackage{soul}
\usepackage{url}
\usepackage[hidelinks]{hyperref}
\usepackage[utf8]{inputenc}
\usepackage[small]{caption}
\usepackage{graphicx}
\usepackage{subcaption}
\usepackage{amsmath}
\usepackage{amsthm}
\usepackage{booktabs}
\usepackage{algorithm}
\usepackage{algorithmic}
\usepackage{todonotes}
\usepackage[switch]{lineno}

\title{Shaping Shared Languages: Human and Large Language Models' Inductive Biases in Emergent Communication}

\author{
Tom Kouwenhoven$^1$
\and
Max Peeperkorn$^2$\and
Roy de Kleijn$^{3}$\And
Tessa Verhoef$^1$\\
\affiliations
$^1$Leiden Institute of Advanced Computer Science, Leiden University, Netherlands,\\
$^2$School of Computing, University of Kent, United Kingdom\\
$^3$Cognitive Psychology Unit, Institute of Psychology, Leiden, the Netherlands\\
\emails
\{kouwenhovent, verhoeft\}@liacs.leidenuniv.nl
}

\newcommand{\citet}[1]{\citeauthor{#1}~\shortcite{#1}}

\makeatletter
\newcommand\citep[2][]{%
    \def\firstcitation{1}%
    [\ifx#1\empty\else#1, \fi%
    \@for\item:=#2\do{%
        \ifnum\firstcitation=1\citeauthor{\item}~\citeyear{\item}%
        \else; \citeauthor{\item}~\citeyear{\item}\fi%
    \def\firstcitation{0}}]
}
\makeatother

\usepackage{newfloat}
\usepackage{listings}
\DeclareFloatingEnvironment[%
    listname={List of Prompts},
    name=Prompt,
    placement=htb]{prompt}

\begin{document}

\maketitle

\begin{abstract}
  Languages are shaped by the inductive biases of their users. 
  Using a classical referential game, we investigate how artificial languages evolve when optimised for inductive biases in humans and large language models (LLMs) via Human-Human, LLM-LLM and Human-LLM experiments. 
  We show that referentially grounded vocabularies emerge that enable reliable communication in all conditions, even when humans \textit{and} LLMs collaborate.
  Comparisons between conditions reveal that languages optimised for LLMs subtly differ from those optimised for humans.
  Interestingly, interactions between humans and LLMs alleviate these differences and result in vocabularies more human-like than LLM-like.
  These findings advance our understanding of the role inductive biases in LLMs play in the dynamic nature of human language and contribute to maintaining alignment in human and machine communication.
  In particular, our work underscores the need to think of new LLM training methods that include human interaction and shows that using communicative success as a reward signal can be a fruitful, novel direction.
\end{abstract}

\section{Introduction}
Languages adapt to how they are learned and used. The primary reason is the continuous influence of individuals' (learning) biases and pressures that slowly shape languages to become more structured, easier to learn and communicatively efficient \cite{Smith2022HowStructure}. Although a wealth of experiments in the field of language evolution have contributed to this \citep[i.a.]{Kirby2014IteratedLearning,raviv2019compositional}, only relatively recently have we started investigating whether these principles can be applied to large language models as well \cite{Galke2023WhatMakes,kouwenhoven-etal-2025-searching}. For instance, more systematic and structured languages are typically easier to learn by humans who are asked to learn novel artificial languages \cite{raviv2021easytolearn}. Recent work by \citet{Galke2023WhatMakes} showed that the same is true for recurrent neural networks and LLMs. Moreover, transmission of initially unstructured language systems over generations of human learners (i.e., iterated learning) increases structure and learnability in these languages \cite{Kirby2015CompressionStructure}. To investigate if LLMs show similar outcomes, \citet{kouwenhoven-etal-2025-searching} created a setting in which LLMs learned an initially holistic unstructured artificial language and then used this repeatedly to communicate in a referential game. They showed that the linguistic structure of these languages increased and more successful communication between LLM agents---again mirroring observations from human experiments \cite{Kirby2015CompressionStructure}.

With AI systems being increasingly incorporated into our daily lives \cite{brinkmann2023machine}, repeated interactions with machines become increasingly important to maintain alignment \cite{mikolov2018roadmap,beuls-van-eecke-2024-humans} and referential grounding \cite{kouwenhoven2022emerging}. 
%
In the case of humans, these repeated interactions cause languages to evolve such that they accommodate for specific abilities and preferences of minority individuals at the group level \cite{josserand2024adapting}. 
Since the seemingly similar ways that languages adapt and optimise as a result of learning and use in Human-Human and LLM-LLM interactions, the question arises whether these processes can also be used to evolve a language that is optimised for humans \textit{and} LLMs. 
In other words; can humans and LLMs collaboratively shape a language that is easy to learn for both, and allows for successful communication? If so, what do these languages look like? 

\begin{figure*}[t!]
  \includegraphics[width=\linewidth]{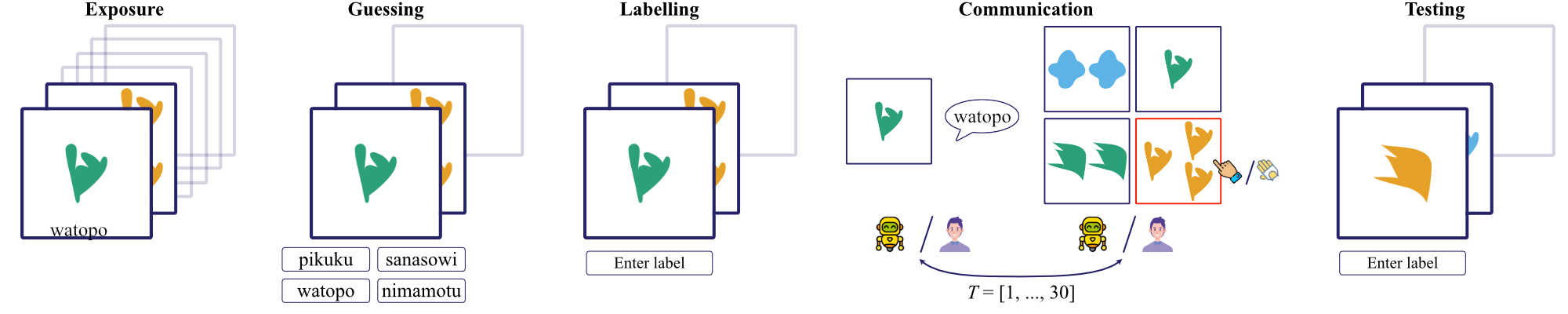}
  \caption{The experimental blocks in our experiment. Participants go through the exposure and guessing block twice before labelling each of the 15 training stimuli in the labelling block. The communication block is performed for 4 rounds each consisting of 30 tasks $T$, where participants alternate speaker-listener roles for each stimulus once. Participants label 27 (15 original and 12 novel) stimuli in the testing block. Image is adopted from \protect\citet{kouwenhoven-etal-2025-searching}.}
  \label{fig:blocks}
\end{figure*}

This work investigates these questions by extending earlier work by \citet{kouwenhoven-etal-2025-searching}. Firstly, we provide experimental data of humans playing the same referential game that \citet{kouwenhoven-etal-2025-searching} used with LLMs, allowing comparisons between languages evolved through LLM-LLM interactions with those resulting from Human-Human interactions. Secondly, we experiment with human-LLM collaboration (\autoref{fig:blocks})\footnote{This study was approved by the ethics department of Leiden University (2024-03-11-R.E. de Kleijn-V1-5354). The code and data are available on OSF \url{https://osf.io/ep6mw/}}. While it is unclear how human and LLM abilities exactly differ, this allows us to test whether an artificial holistic language can be optimised for the inductive biases of two different types of language learners. If shared vocabularies of signals and meanings emerge that allow for successful communication, one could argue that there has been some form of referential grounding, a prerequisite for successful communication \cite{clark1991grounding}. Finally, Human-LLM collaboration allows investigating if and how the evolved language differs from languages that evolve within Human-Human and LLM-LLM interactions. 

Our results show that structured and referentially grounded languages can emerge when humans and LLMs interact repeatedly. The languages that emerge from these interactions tend to be more human-like than LLM-like, suggesting that the LLMs are flexible towards the strong human preferences that shape the languages. Finally, languages optimised for LLMs have less variation and are more degenerate than those optimised for humans. 

\section{Background}
We discuss relevant research on language evolution, the role of inductive biases, and their relevance to LLM research.

\subsection{Language evolution}
Language allows us to communicate successfully because of the vocabulary we share, but also due to its open-ended nature, enabling the possibility to express novel meanings via compositional semantics. This defining feature of human language means that the meaning of any phrase is derived from the meanings of its individual components and the rules by which they are combined \cite{hockett1960origin}. The evolution of compositionality has been investigated abundantly in the field of language evolution through human experiments \citep[e.g.]{kirby2008cumulative,raviv2019compositional} and computational simulations \citep[e.g.]{deBoer2006computer,steels2012grounded,lazaridou2020emergent,kouwenhoven-etal-2024-curious}. These experiments typically involve learning artificial languages or playing a signalling game. 
Here, \textit{learning} artificial languages imposes a constraint which is believed to lead to more structured languages \cite{Kirby2015CompressionStructure}. \textit{Communication} in signalling games imposes a pressure for expressivity, requiring participants to develop a vocabulary of signal-meaning mappings that allows them to communicate about novel stimuli. In this case, some form of referential grounding must be established through the process of repeated interactions. Participants--human or machine--generally establish novel signal-meaning mappings quickly that enable successful communication. 

\subsection{Inductive biases}\label{sec:biases}
An important aspect in this paper is the notion of biases. Here, we do not so much focus on behavioural biases observed in humans (e.g., the confirmation bias), but are rather interested in implicit inductive biases that may result in biased language learning. 
This is relevant since seemingly arbitrary aspects of linguistic structure may actually result from general learning and processing biases deriving from the structure of thought processes, perceptuo-motor factors, cognitive limitations, and pragmatics \cite{Christiansen2008LanguageBrain}. 
For humans, inductive biases like a preference for compressibility, simplicity, or efficiency \cite{Kirby2015CompressionStructure,tamariz2015culture,gibson2019howefficiency} naturally shape how languages evolve. Some even argue that humans' cognitive limitations may be \emph{beneficial} for language acquisition \cite{decaro2008individual,poletiek2018cogsci}. 
In the case of simulations of (reinforcement learning) agents, inductive biases typically do not match those present in humans. As such, they are often induced artificially by incorporating biases to guide learning dynamics as a means to recover human-like properties \citep[for a review see]{galke2024emergentcommunicationlearningpressures}. 
In the case of LLMs, being fundamentally different from humans (although this is an ongoing debate \citet{kozachkov2023pnas}), we focus on increasingly apparent inductive biases of the transformer architecture \cite{futrell2025linguisticslearnedstopworrying} that may influence how languages evolve in the context of our experiment. 

One example is a bias for simplicity. 
\citet{rende2024a} carefully cloned training data such that texts only contained between-token interactions up to a certain degree. Revealing that transformers first learn low-degree between-token interactions, and only later learn high-degree interactions.
Similarly, LLMs pick up grammar as the simplest explanation for data early during training. Only shortly thereafter, general linguistic capabilities arise \cite{chen2024sudden}. 
Moreover, transformers have an inductive bias favouring structure in (natural) language. For example, GPT-2 models struggle to learn impossible languages (e.g., languages lacking hierarchical structure or having unnatural or irreversible word orders) compared to English \cite{kallini-etal-2024-mission}, indicating that structure aids language learning. Additionally, the ability to generalise to novel stimuli increases when LLMs learn from more structured artificial languages \cite{Galke2023WhatMakes,kouwenhoven-etal-2025-searching}. 
Recent work also revealed a primacy and recency bias in LLMs. They handle information better when it appears either at the beginning or towards the end of a prompt \cite{liu-etal-2024-lost,mina-etal-2025-cognitive}. 
Finally, LLMs have an inductive preference for verbose answers \cite{zheng2023verbosity,saito2023verbosity}, while humans prefer short, efficient answers \cite{gibson2019howefficiency}.

Although the underlying mechanisms of these biases differ between humans and machines, we find substantial overlap in terms of their behavioural effects. As such, we hypothesise that the aforementioned effects of continuous learning and use of language will also come into play when humans and machines collaborate and result in a language optimised for the preferences of both entities.

\subsection{Why is this relevant for LLMs?}
It is increasingly assumed that LLMs can be used as models of language \cite{milliere2024languagemodelsmodelslanguage} and that classical approaches from emergent communication can inform more human-like language learning in machines \cite{beuls-van-eecke-2024-humans,galke2024emergentcommunicationlearningpressures}. Moreover, language modelling and linguistics should complement each other \cite{futrell2025linguisticslearnedstopworrying} as comparing LLMs to human language users, can help answer cognitive and typological questions \cite{warstadt2022artificial,van-dijk-etal-2023-large}. Vice-versa, methods from psychology can help to quantify inductive biases of LLMs \cite{Griffiths2024Bayes,galke2024emergentcommunicationlearningpressures} or vision-and-language models \citep[e.g.]{verhoef-etal-2024-kiki} and compare them to known biases in humans.

For example, the process of iterated learning, in which the transmitted information will ultimately come to mirror the minds of the learners \cite{griffiths2007language}, has been used to discover inherent LLM biases. \citet{ren2024bias} for example showed that iterated learning causes subtle biases in LLM priors to be gradually amplified, \citet{kouwenhoven-etal-2025-searching} concluded that artificial languages can be optimised for LLM-augmented agents with iterated learning, and \citet{shumailov2024ai} argue that generative models converge on uninterpretable junk when they are trained on AI-generated data. While the latter is typically seen as drift, crucially, we argue that what this shows is that the generated content is slowly shaped to be optimised for model preferences, \textit{not} for humans. To prevent what \citet{shumailov2024ai} call model collapse, they argue that genuine human interactions with systems will be increasingly important. Similarly, \citet{smith2024ai} responded that, like in human language transmission, the need to be expressive may prevent both the convergence on a few frequent uninformative sentences and the emergence of a long tail of uninterpretable junk.


These findings advance our understanding of internal LLM representations. Thereby contributing to maintaining alignment and mutual understandability between humans and machines in interaction. 
We address this by examining how adaptation processes unfold when humans and machines interact and develop a novel artificial language together.

\section{Methodology}
This experiment revolves around the classical referential Lewis game as implemented in \citet{kouwenhoven-etal-2025-searching}, who based their setup on previous work in emergent communication\citep[e.g.]{raviv2021easytolearn,Kirby2015CompressionStructure}. We extend the setup used by \citet{kouwenhoven-etal-2025-searching} to incorporate humans. In total, 45 participants participated in the experiment, 30 of whom formed 15 Human-Human pairs, and the remaining 15 interacted with an LLM in a Human-LLM setup. This allows us to directly compare languages adapted for human preferences to those adapted for the LLM-LLM simulations \cite{kouwenhoven-etal-2025-searching}. But perhaps most interestingly, the Human-LLM condition provides an opportunity to investigate whether languages can be optimised for entities with different mechanisms and cognitive capacities (e.g., memory), and if so, we can see what these look like.

During the experiment, participants learn an artificial language and use it during communication. This language comprises a meaning space consisting of three attributes (shape, colour, and amount) that each can have three values, totalling to 27 unique stimuli. The corresponding labels are initialised following the design of \citet{kirby2008cumulative}, creating a holistic artificial language without structure (e.g., "watopo", "sanasowi", "pikuku") that contained a limited set of characters to prevent participants from writing English words. Participants first learn 15 random signal-meaning pairs individually through exposure, guessing, and labelling blocks. Hereafter, participants use the newly acquired language to communicate in a referential game. In this game, the speaker observes a target stimulus and labels it. Using this label as a signal, the listener is then tasked with identifying the correct target among three distractors. Cooperation is successful when the listeners' guess matches the target. After the communication block, there is a testing block in which participants individually label 27 meanings, including 12 unseen meanings. The duration of the entire experiment is roughly 70 minutes. An experiment overview is provided in \autoref{fig:blocks}.

\subsection{LLMs as participants}

\begin{figure}[t!]
  \includegraphics[width=\linewidth]{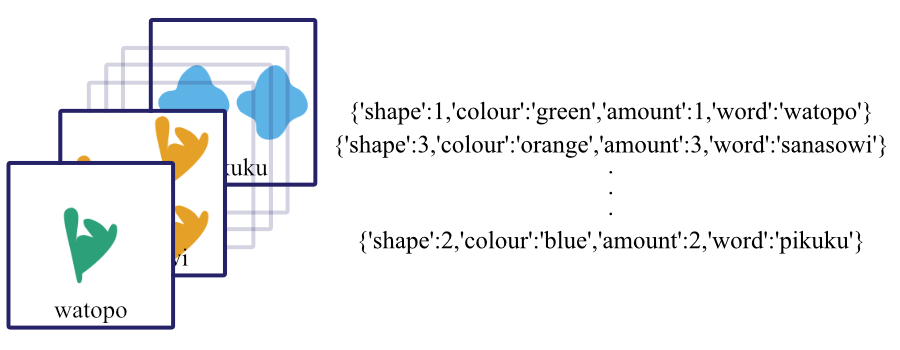}
  \caption{Left: Humans learn the language by being exposed to stimuli and the corresponding signals in the exposure block. Right: LLMs learn the same vocabulary by virtue of in-context learning. A JSON-like structure containing the signal-meaning mappings is prepended to each prompt to serve as learning stimuli.}
  \label{fig:learning}
  \vspace{-0.75em}
\end{figure}

\begin{prompt*}[ht!]
\scriptsize
\begin{lstlisting}[mathescape=true,escapeinside={*@}{@*}]
<|begin_of_text|><|start_header_id|>system<|end_header_id|> You are a language learner who has to learn an artificial language with words and their corresponding features. Your task is to complete the vocabulary by generating a word that describes the last item. Only respond with the word.<|eot_id|><|start_header_id|>user<|end_header_id|>\n
{'shape':2,'colour':'orange','amount':1,'word':'giniwite'}
    :
{'shape':3,'colour':'blue','amount':2,'word':'tusetetu'}
{'shape':1,'colour':'green','amount':3,'word':'<|eot_id|><|start_header_id|>assistant<|end_header_id|>[comp/prefill]
\end{lstlisting}
\caption{A prompt snippet used for labelling and guessing. During communication, we add a \texttt{communicativeSuccess} attribute, update the system prompt to inform about the communicative task, and instruct that `Communicative success is important'.}
\label{fig:full-cmpl-prompt-labelling}
\end{prompt*}

Human participants learn the language by going through exposure and guessing blocks twice. They iteratively go over the 15 training stimuli and extract some apparent, but not present, patterns or consistencies. They are then tasked to label the stimuli, before moving on to the communication block in which they interact with another human or a LLM. In either case, they were told that they interacted with a human. The LLM agents, however, are not updated and receive instructions to learn the languages by virtue of in-context learning. 
Specifically, the stimuli are presented in a structured JSON-like format (\autoref{fig:learning}) that proved effective in \citet{Galke2023WhatMakes} and \citet{kouwenhoven-etal-2025-searching}. As such, we assume that these signal-meaning mappings in the context of a prompt provide enough (distributional) information for a LLM to learn a mapping between the attributes of the stimuli and signal syllables (\autoref{fig:full-cmpl-prompt-labelling}). Although the prompt structure `invites' the LLM to infer a signal from the stimulus attributes, we are agnostic about how exactly and what kind of mapping is deduced, but are interested in the resulting behaviours. 
In the experiments, we use the instruction-tuned variant of Llama 3 70B \cite{dubey2024llama3herdmodels} with greedy sampling. We opt for an instruction-tuned model since this allows us to specify the need for communicative success. This potentially affects how the model's inductive biases are expressed, but we leave this to future work.

One of essentially two tasks is performed throughout the experiment: labelling or guessing. The labelling block and speaking in the communication block involve labelling, and the guessing block and discrimination during communication involve guessing. Generating signals is achieved through prompt completion. Guessing is done by prefilling the prompt with distractor stimuli or labels and selecting the item with the highest probability. This alleviates LLMs' inconsistent behaviour in answering multiple-choice questions \cite{khatun2024study}, and follows recommendations from computational linguistics \cite{hendrycks2021measuring,wang-etal-2024-answer-c}. During communication, we add a \texttt{communicativeSuccess} attribute set to $1$ if the previous interaction for this stimulus was successful and zero otherwise. This attribute functions as a memory between interactions and acts as a pressure for expressivity. In human language evolution, such pressure plays an important role since it prevents languages from becoming degenerate \cite{Kirby2015CompressionStructure}. Importantly, the agents observe the training vocabulary in their context \textit{with} the current stimulus in the guessing and labelling block, rendering them as simple look-up tasks. We do, however, \textit{not} include the current stimulus during communication and testing, requiring the agents to extract an appropriate mapping and generalise to new stimuli. Akin to common practice in older simulations \citep[e.g.]{steels2012grounded}), the agent vocabularies are updated when labels are generated after the labelling block and during the communication block. In particular, each time a label is generated for a stimulus, it replaces its current label value. Hence, the agent's vocabularies of signal-meaning mappings evolve during the simulation.
As such, prompts are slightly different after each interaction. Moreover, given the primacy and recency bias in LLMs \cite{liu-etal-2024-lost,mina-etal-2025-cognitive}, we shuffle the vocabulary before creating prompts to account for unwanted ordering effects.  


\subsection{Metrics}
Besides comparing the percentage of communicative success (\textit{PercCom}), the primary goal of this work is to understand what a language looks like when optimised for different entities. Specifically, we investigate whether the languages display some degree of structure in the form of compositionality. In this experiment, this means that attribute values are denoted with label parts that are reused to describe other similar stimuli. Capturing this is not at all trivial, especially provided the freedom given to participants when they label stimuli. A common metric that gauges whether similar meanings map to similar signals is Topographic Similarity \citep[\textit{TopSim}]{Henry2006Understanding}. While providing a good indication of compositional language use, it does not take variability in language, such as word-order freedom, into account. It could therefore show an incomplete picture (i.e., a low \textit{TopSim}) since languages can still be compositional despite having multiple word orders, the existence of synonyms, or homonyms \cite{conklin2023compositionality}. Hence, we report multiple metrics in addition to \textit{TopSim} that together indicate the degree of compositionality. Specifically, on \textit{synonymy} (one-to-many mappings), \textit{homonymy} (many-to-one), and word order freedom (\textit{Freedom}), for which \citet{conklin2023compositionality} proposed entropy-based metrics. A language where each attribute value is encoded by a single character in a position has low entropy, and thus a low \textit{synonymy}. Languages with a uniform distribution over all characters to refer to an attribute value have high synonymy. \textit{Homonymy} is similar, it looks at how many attribute-values a character in a position can refer to, i.e., when $\textit{homonymy} \approx 1$ characters can map to multiple attribute-values. Finally, we compute word-order freedom (\textit{freedom}) to account for variability in the order by which labels are composed. It assesses whether each value of a specific attribute is encoded in a specific position of the label, i.e., there is little freedom, or whether attribute values can be encoded in any position of the label, i.e., displaying a high degree of word order freedom ($\textit{Freedom} \approx 1$).

Systematic generalisation to novel stimuli is assessed through the generalisation score \textit{GenScore} from \cite{raviv2021easytolearn}. It gauges whether the labels produced for unknown (i.e., testing) stimuli are labelled in consistent ways with labels produced for similar known (i.e., training) stimuli. 
In addition to character-based metrics, we assess whether participants re-use parts of labels in different labels by computing the \textit{Ngram} diversity \cite{li-etal-2016} over all the produced labels in a block. \textit{Ngram} diversity is the average ratio of unique vs. total \textit{Ngrams} for $N \in \{1, 2, 3, 4\}$ in all labels. Low \textit{Ngram} diversity implies that labels are composed of reused parts and high diversity means that labels do not share many \textit{Ngrams}, thus are very different. 
The percentage of unique labels captures the degree of degeneracy (\textit{RatioUniLabels}). Finally, we measure whether a pressure for communicative success, known to drive efficiency in human experiments \cite{smith2020communicative}, results in shorter labels using \textit{WordLength}. 

  
\section{Evaluation}
We use linear mixed effect models to analyse our results. Specifically, we fit $PercCom \sim Metric + (1|RoundId)$ where $Metric$ can either be \textit{TopSim}, \textit{RatioUniLabels} and is the average value of two players in a round. To measure effects across conditions, we use $PercCom \sim Metric + Metric*Condition + (1|RoundId)$. The slope $\smash{\hat{\beta}}$ determines the direction of the effect and the rate of change. Additionally, we use conditional $R^2_c$, and marginal $R^2_{m}$ \cite{nakagawa2013r2}. The former considers fixed and random effects to show how much variance can be explained by the model. Higher values of $R^2_c$ indicate that the model captures more variance and that correlations are stronger. $R^2_{m}$ describes how much variance can be explained by the fixed effects. We report Pearson's R to describe the relationship between \textit{TopSim} and \textit{GenScore}, and use a paired T-test, or Welch's test when assumptions on normality and variance were not met, to assess whether metrics significantly differ.  

\section{Results}
\begin{figure}
  \includegraphics[width=\linewidth]{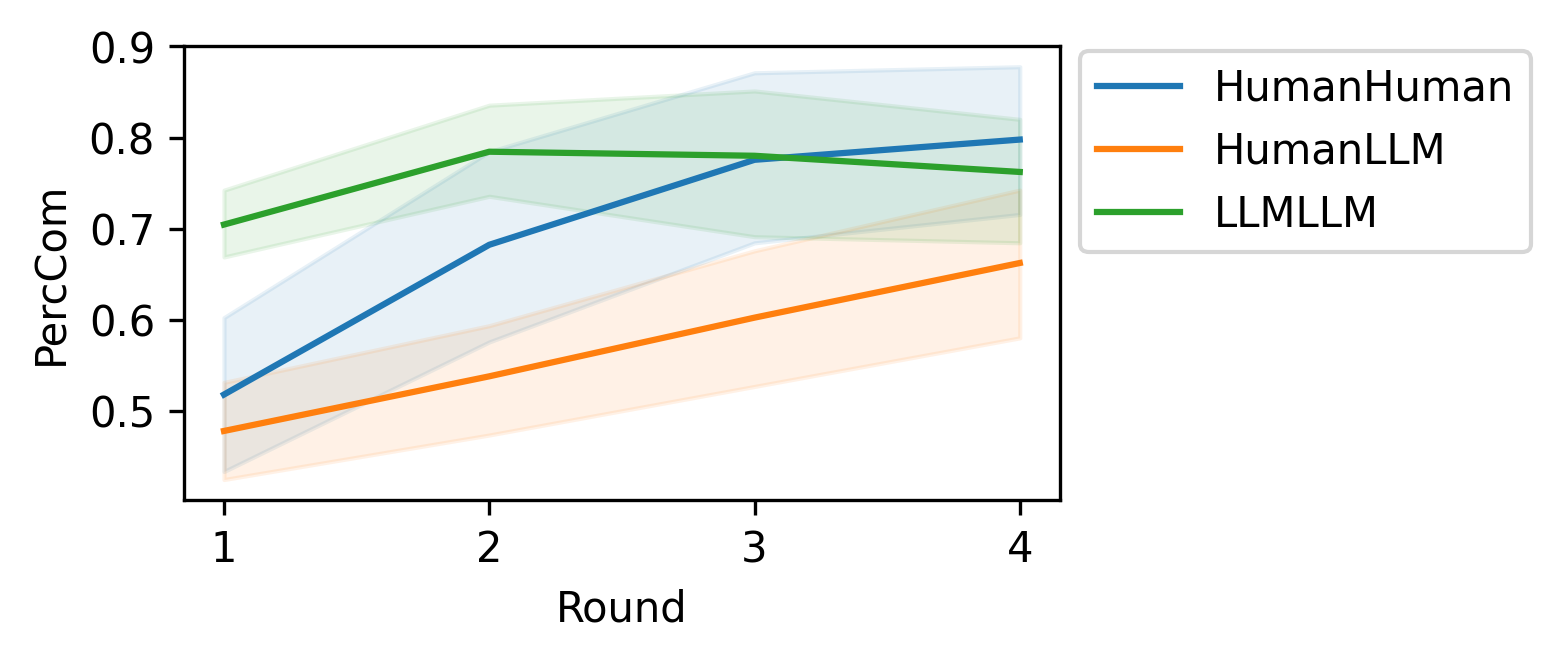}
  \caption{The average communicative performance (\textit{PecrCom}) per round across the conditions. Communication steadily increases over rounds except for the LLM-LLM condition, in which coordination happens in the first round but does not increase afterwards.}
  \label{fig:communicativesuccess}
\end{figure}

\paragraph{Human artificial language learning} happened in a way that was expected based on earlier work \cite{Kirby2015CompressionStructure,raviv2021easytolearn}. The results of these 15 Human-Human ($n=30$) experiments act as a benchmark of human behaviour in our setup. We find that learning artificial languages is not a trivial task. After two rounds of exposure, labels were correctly guessed approximately half the time ($47.0\%\pm49.9$). Freely labelling stimuli was done correctly only in $10.4\%\pm30.6$ of the labels. 
Nevertheless, reliable communication protocols emerged during the communicating block, interactions were significantly more successful in the final round compared to the first round (\autoref{fig:communicativesuccess}, $t(14)=-6.30, p<.001, d=1.63, PercCom_{r1}=.518\pm.176, PercCom_{r4}=.798\pm.169$). \textit{TopSim} positively influenced communication (\textit{PercCom}, $\hat{\beta}=.087\pm.009, R^2_{c}=.731, R^2_{m}=.714, p<.001$) and generalisation to new stimuli was more consistent when the languages that evolved during communication displayed more \textit{TopSim} ($r=.826, p<.001$).
A qualitative inspection revealed that, during communication, participants quickly replaced the labels they learned before, with only parts of labels `surviving' this cut. Moreover, the number of shapes displayed (1, 2, or 3) was sometimes encoded by repeating the shape and colour labels several times, e.g., "pufepufe" was used to indicate two green shapes. Although expressive, this solution does not generalise to larger numbers and is therefore arguably not compositional. 

\begin{figure*}[t!]
  \includegraphics[width=\linewidth]{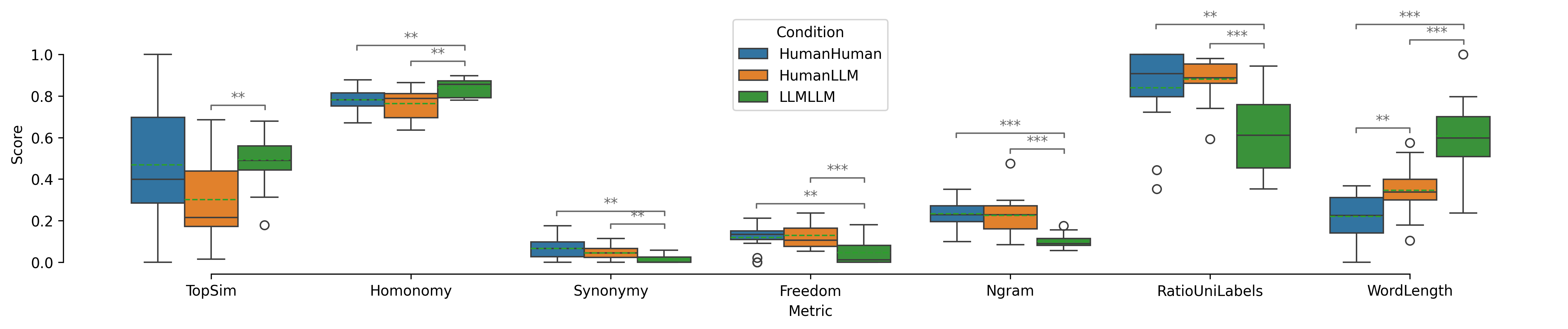}
  \caption{An overview of structure metrics used to measure compositional structure in the languages produced in the testing block. Generally, languages optimised for LLMs differ from those optimised for humans. Languages optimised for both mediate these differences. The asterisks indicate whether an independent Welch's t-test reveals a significant difference between the conditions where $\ast\ p < .05$, $\ast\!\!\ast\ p < .01$, $\ast\!\!\ast\!\!\ast\ p < .001$. \textit{TopSim} and \textit{WordLength} are normalised to values between 0 and 1 for visualisation purposes.}
  \label{fig:combinedmetrics}
\end{figure*}

\paragraph{Artificial language learning in LLMs}\label{test} was assessed by \citet{kouwenhoven-etal-2025-searching}. Here, we briefly discuss the results of LLM-LLM ($n=15$) simulations \footnote{We also ran simulations with other (smaller) models (i.e. Llama 3 8B, OLMo 2 7B 1124 \& OLMo 2 13B 1124). Communication was successful but more difficult. OLMo 2 models struggled most.}. LLMs correctly guessed labels for $97.3\%\pm16.1$, and the produced labels in the labelling block exactly matched the initial labels for $45.3\%\pm.498$. This indicates that learning the language is easier for LLMs than for humans. This is not surprising since the target stimulus is present in the prompt context in these blocks and there is virtually no memory constraint. Communication happens reliably as well (\autoref{fig:communicativesuccess}), however, communication can--but does not always--result in degenerate vocabularies with few uniquely used labels ($RatioUniLabels=.621\pm.198$), significantly differing from human diversity in labels ($RatioUniLabels=.841\pm.201, t(28)=-3.01, p=.005$, \autoref{fig:combinedmetrics}). Interestingly, this happens even though the ratio of unique labels is modestly related to \textit{PercCom} during communication ($\smash{\hat{\beta}}=.300\pm.134, R^2_{c}=.180, R^2_{m}=.092, p=.025$), suggesting that expressiveness is beneficial. A tentative explanation could be that aligning vocabularies happens much faster than in the other conditions. While humans would optimise languages whilst retaining expressiveness, LLMs start producing more duplicate and longer labels.

\paragraph{What about Human-LLM communication?} Our main contribution comes from the Human-LLM condition in which participants ($n=15$) collaborated with LLMs. Successful communication necessitates that both entities adopt their behaviours and that a reliable referentially grounded vocabulary emerges. This is especially interesting since we observed that the abilities to learn artificial languages differ between humans and LLMs, and that the optimised languages differ in their use of homonyms, i.e., duplicate labels for different meanings. Despite these differences, communication \textit{is} still possible ($PercCom=0.662\pm0.161$, \autoref{fig:communicativesuccess}). The final performance was lower than the other conditions, but the data suggests that prolonged interactions may result in higher communicative success. This exciting result shows that, even though the learning mechanisms of both entities may initially learn different signal-meaning relations, communication is possible. As such, the process of repeated learning and use of these artificial languages can overcome initial differences and indeed shape languages to be optimised for humans \textit{and} LLMs. Out of 15 participants, 9 believed their partner was another human, despite communicating with an LLM. Performance did not change significantly as a result of this.

\paragraph{What do these languages look like?}
Having established that participants can communicate, we now examine if and how languages differ across different experimental conditions, focusing on the language metrics. 
Generally, we find that languages optimised for LLMs differ from those optimised for humans, and that languages optimised for \textit{both} are more human-like than LLM-like (\autoref{fig:combinedmetrics}). This is most notably visible in the ratio of uniquely produced labels, their respective lengths, and the \textit{Ngram} diversity. 
Languages optimised for humans contain more unique labels, have higher \textit{Ngram} diversity, and are shorter when compared to languages optimised for LLMs. The latter of which corroborates the well-known human preference for efficient communication \cite{smith2020communicative}. 
The length, number of unique labels, and diversity of label parts resulting from human-LLM collaboration seem to adhere more to human preferences than to LLM preferences. A similar pattern is visible for the compositionality metrics that allow for variation. There is more \textit{homonymy} in LLM-optimised languages than in human-LLM languages, suggesting that the meanings of these words should be disambiguated by the context (i.e., the distractor stimuli) in which they appeared. This strategy is not straightforward and perhaps requires more cognitive capacity than available for humans, and may therefore be lower in the collaborative condition. It moreover seems that humans introduce synonymy into the languages, i.e., they use more than a single character to refer to specific attribute values. This introduces variability that can explain why the ratio of unique labels and \textit{Ngram} diversity is higher in the collaborative condition. Finally, the word order of messages is somewhat flexible for humans, while LLMs converge to a more fixed word order. The languages shaped by both seem to have human-like word order freedom. While these variations may introduce difficulties for LLMs to decode the meanings during communication, we do not find that \textit{PercCom} is affected by the degree of \textit{homonymy}, \textit{synonymy}, or \textit{freedom}.

The canonical \textit{TopSim} metric suggests structure is lower when humans and LLMs collaborate compared to LLM simulations. This makes intuitive sense, given our observation that it was also more difficult to establish successful communication. Nevertheless, a linear mixed effects model fitted to predict \textit{PercCom} with \textit{TopSim}, the experimental condition, the interaction between them, and a random effect for round revealed that \textit{TopSim} strongly affects \textit{PercCom} ($\hat{\beta}=.092\pm.008, R^2_{c}=.580, R^2_{m}=.580, p<.001$). This means that irrespective of the experimental condition, a higher degree of structure in produced labels was beneficial for communication. Moreover, generalisation to novel stimuli happened more consistently when the languages in the last round of communication displayed more structure. Again, confirmed by the mixed effects model predicting \textit{GenScore} with \textit{TopSim}, the condition, their interaction and round as random effect ($\hat{\beta}=.047\pm.006, R^2_{c}=.784, R^2_{m}=.673, p<.001$).

Generally, we find that while remarkably human-like, the languages shaped by intrinsic LLM constraints are in fact subtly different from those shaped by humans. Thereby providing a more nuanced view of earlier work \cite{kouwenhoven-etal-2025-searching} that only looked at \textit{TopSim} and \textit{NGram} diversity. Returning to the question of what languages optimised for entities with different inductive biases look like; they seem to be shaped in such a way as to conform more to human pressures than those present in LLMs.





\section{Discussion}
The primary goal of our work was to investigate if and how artificial languages differ when optimised for human \textit{and} artificially intelligent language users.
We extended work suggesting that LLMs can shape and use languages in referential communication \cite{kouwenhoven-etal-2025-searching}.
Their setup was adapted so that participants could interact with other human participants and with LLMs.
This enabled controlled comparisons between the languages that evolve under different conditions. Our findings showed that human pairs, LLM pairs and Human-LLM pairs can learn and successfully use languages in a referential game. This suggests that mechanisms that influence how language evolves for humans, specifically, learning and using a language repeatedly \citep[e.g.]{Smith2022HowStructure}, also apply to computational and collaborative Human-LLM settings. In all conditions, successful communication was achieved by optimising an initially holistic unstructured vocabulary, to fit better with the inductive biases of the language users. Comparison of the languages across conditions revealed that 1) while very human-like, LLM languages tend to be more strict (i.e., there is little variation) and that 2) languages adapted for human-LLM pairs tend to be more human-like than LLM-like (i.e., they are more diverse and have variation). Overall, our findings corroborate earlier claims that interactions between humans and machines are beneficial to establishing some form of referential grounding \cite{mikolov2018roadmap,kouwenhoven2022emerging,beuls-van-eecke-2024-humans}. 

On the level of vocabulary, the ratio of uniquely produced labels by LLMs revealed that vocabularies can become degenerate. While this is also observed in human experiments when there is no pressure against it \cite{kirby2008cumulative}, communicative success as a pressure is typically enough to prevent this \citep[e.g.]{Kirby2015CompressionStructure}. In contrast, even though the LLMs in our experiments were tested in a communicative setting, this did not prevent the languages from becoming underspecified. Possibly because the instruction to obtain communicative success was not explicit enough and did not induce enough pressure for expressivity. Or it may be that the distractors did not require the labels to be very specific but instead allowed using underlying concurrences that were picked up by LLMs but not by humans. 
This would also explain the wide range of scores on this \textit{RatioUniLabels}. On a character level, we see related patterns in the form of high levels of \textit{homonymy}, meaning that attribute values could be associated with multiple label characters, and that context was necessary to disentangle the correct meaning. While the duplicate labels can explain these scores for LLM simulations, this is not the case for humans. Here, the surprising behaviour of repeating label parts to indicate the \textit{amount} attribute can explain the \textit{homonymy} values. 


Since we do not modify the tokeniser to deal with our artificial languages, the tokenisation process plays an important role in these simulations.
One could argue that this helps in learning a mapping between tokens and meanings. 
The meaning attributes and their values are common English words, while the initialised artificial languages are non-words tokenised into separate tokens. 
Meaning that the LLM is presented with a parsed set of attribute meanings and chunks of labels (i.e., the tokens). 
All that is left is to attend to a specific token given a specific meaning attribute, which is exactly what a transformer model is made for. 
This does, however, not undermine that these models \textit{indeed} attribute attention correctly and that this produces human-like languages. 

The fact that a shared referential communicative system can be established is quite remarkable. Humans and LLMs may well rely on completely different mechanisms, and learn different relations between meanings and signals. Yet, their vocabularies become referentially grounded and are pragmatically understood by both humans and LLMs. This confirms that even though LLMs are not trained for this task, they can be used as relatively unbiased language learners \cite{wilcox2023using}, providing a concrete example of how a pragmatic view of understanding as argued for by \citet{van-dijk-etal-2023-large}, can be beneficial for collaborative tasks. This work also underscores the point made by \citet{milliere2024anthropocentric} that how LLMs or other AI models solve a cognitive task cannot be used as an argument against particular cognitive competences or language understanding, as long as the solution generalises.

Our results concretely corroborate the idea that insights from emergent communication literature can inform and improve language learning in language models \cite{smith2024ai,beuls-van-eecke-2024-humans,galke2024emergentcommunicationlearningpressures}. We observed that just as languages accommodate for specific abilities and preferences in humans \cite{josserand2024adapting}, Human-LLM languages also adapt to the abilities and preferences of their users in that they are more human-like than LLM-like. 
Specifically, human preferences for simplicity and efficiency \cite{Kirby2015CompressionStructure,gibson2019howefficiency} likely drove vocabulary diversity while reducing lengths to human-like levels.
This indicates that, in this experiment, LLMs are more flexible communicative partners than humans. 
These findings reinforce the idea that repeated interactions with humans are crucial to maintaining referentially grounded human-like vocabularies instead of training only on recursively generated data \citep[e.g.]{shumailov2024ai} or using AI-augmented optimisation algorithms \citep[e.g.]{lee2024rlaif}. 
In particular, this underscores the need for new LLM training methods that include human interaction and shows that using communicative success as a reward signal can be a fruitful, novel direction \citep[e.g.]{stöpler2025developmentallyplausiblerewardscommunicative}


Finally, we acknowledge that our results depend on methodological considerations such as in-context learning, the prompt format, and the sampling method. 
However, the primary goal was to extend previous work by investigating if languages optimised for human \textit{and} LLM preferences can evolve. 
As such, we stayed close to well-known experimental methods in language evolution research and used prompts developed in \citet{kouwenhoven-etal-2025-searching}. Importantly, we did not optimise for communicative success, human-like results, or compositional vocabularies. 

\section{Conclusion}
Given the growing presence of contemporary LLMs in everyday life, there is an increasing need to understand their inductive biases to maintain alignment with humans. 
We tested whether general mechanisms of language learning and use have similar effects in an artificial language learning experiment conducted with Human-Human, LLM-LLM, and Human-LLM pairs.
We show that referentially grounded vocabularies emerge in all conditions, indicating that initially unstructured artificial languages can be optimised for inductive biases of different language users. 
Comparisons across conditions revealed that, while similar to human vocabularies, LLM languages are subtly different. 
Interestingly, these differences are alleviated when humans and LLMs collaborate. 
This underscores that to achieve successful interactions between humans and machines, it is essential to optimise for communicative success. 
Overall, these findings advance our understanding of how LLMs may adapt to the dynamic nature of human language, contribute to its evolution, and maintain alignment with human understanding of language. 
While our setup only uses simple stimuli and basic languages, achieving this for human-level languages is a key research direction towards more natural language learning in LLMs.

\bibliographystyle{named}
\bibliography{custom}



\end{document}